\documentclass[journal]{IEEEtran}
\usepackage{mathrsfs}

\ifCLASSOPTIONcompsoc
\else
\fi
\ifCLASSINFOpdf
\else
\fi
\usepackage{graphicx}
\usepackage{subfigure}
\usepackage{cite}
\newtheorem{theorem}{Theorem}

\newtheorem{lemma}{Lemma}

\newtheorem{proposition}{Proposition}
\newtheorem{question}{Question}
\newtheorem{definition}{Definition}

\usepackage{graphicx}
\usepackage{subfigure}
\begin{document}

\title{Is extreme learning machine feasible? A theoretical assessment (Part II)}

\author{Shaobo Lin,~Xia Liu,~Jian Fang,
        and~Zongben Xu\thanks{ Corresponding
author: Zongben Xu: zbxu@mail.xjtu.edu.cn}
\IEEEcompsocitemizethanks{\IEEEcompsocthanksitem S. Lin, Xia. Liu,
J. Fang and Z. Xu are  with the Institute for Information and System
Sciences, School of Mathematics and Statistics, Xi'an Jiaotong
University, Xi'an 710049, P R China}}

\IEEEcompsoctitleabstractindextext{%
\begin{abstract}
An extreme learning machine (ELM) can be regarded as a two stage
feed-forward neural network (FNN) learning system which  randomly
assigns the connections with and within hidden neurons in the first
stage and tunes the connections with output neurons in the second
stage. Therefore, ELM training is essentially a linear learning
problem, which significantly reduces the computational burden.
Numerous applications show  that such a computation burden reduction
 does not degrade the generalization capability. It has,
however, been open that whether this is true in  theory. The aim of
our work is to study the theoretical feasibility of ELM by analyzing
the pros and cons of ELM.
 In the previous part on this topic, we pointed
out that via appropriate selection of the activation function, ELM
does not degrade the generalization capability in the expectation
sense. In this paper, we launch the study in a different direction
and show that the randomness of ELM also leads to   certain negative
consequences. On one hand, we find that the randomness causes an
additional uncertainty problem of ELM, both in approximation and
learning. On the other hand, we theoretically justify that there
also exists an activation function such that the corresponding ELM
degrades the generalization capability. In particular, we prove that
the generalization capability of ELM with Gaussian kernel is
essentially worse than that of FNN with Gaussian kernel. To
 facilitate the use of ELM, we also provide a
 remedy to such a  degradation.  We find that
  the well-developed coefficient
regularization technique can essentially improve the generalization
capability. The obtained results reveal the essential characteristic
of ELM and give theoretical guidance concerning how to use ELM.
\end{abstract}


}

\maketitle

\IEEEdisplaynotcompsoctitleabstractindextext

\IEEEpeerreviewmaketitle

\section{Introduction}

An extreme learning machine (ELM) is a feed-forward neural network
(FNN) like learning system whose connections with output neurons are
adjustable, while the connections with and within hidden neurons are
randomly fixed.  ELM then transforms the training of a FNN into a
linear problem in which only connections with output neurons need
adjusting. Thus the well-known generalized inverse technique
\cite{Rao1971,Serre2002} can be directly applied for the solution.
Due to the fast implementation, ELM has been widely used in
regression \cite{Huang2006}, classification \cite{Huang20111}, fast
object recognition \cite{Xu2012}, illuminance prediction
\cite{Ferrari2012}, mill load prediction \cite{Tang2012}, face
recognition \cite{Marques2012} and so on.

Compared with the enormous emergences of applications, the
theoretical feasibility of ELM is, however,   almost vacuum. Up till
now, only the universal approximation property of ELM is analyzed
\cite{Huang2006,Huang20061,Huang20062,Zhang2012}. It is obvious that
one of the main reasons of the low computational burden of ELM is
that only a few neurons are utilized  to synthesize the  estimator.
Without such an attribution, ELM can not outperform other learning
strategies in implementation. For example, as a special case of ELM,
learning in the sample-dependent hypothesis space (the number of
neurons equals to the number of sample)
\cite{Sun2011,Tong2010,Wu2008} can not essentially reduce the
computational complexity. Thus, the universal approximation property
of ELM is too weak and can not capture the essential characteristics
of ELM. Therefore, the generalization capability and approximation
property of ELM should be investigated. The former one focuses on
the relationship between the prediction accuracy and the number of
samples, while the latter one discusses the dependency between the
prediction accuracy and the number of hidden neurons.

The aim of our study is to theoretically verify the feasibility of
ELM by analyzing the pros and cons of ELM. In the first part on this
topic \cite{Liu2013}, we casted the analysis of ELM into the
framework of statistical learning theory and concluded that with
appropriately selected activation functions (polynomial,
Nadaraya-Watson and sigmoid), ELM did not degrade the generalization
capability in the expectation sense. This means that, ELM reduces
the computation burden without sacrificing the prediction accuracy
by selecting appropriate activation function, which can be regarded
as the main advantage of ELM. To give a comprehensive feasibility
analysis of ELM, we should also study the disadvantage of ELM and,
consequently, reveal the essential characteristics of ELM.

Compared with the classical  FNN learning \cite{Hagan1996}, our
study in this paper shows that there are mainly two disadvantages of
ELM. One is that the randomness of ELM causes an additional
uncertainty problem, both in    approximation and learning. The
other is that there also exists a generalization degradation
phenomenon for ELM with inappropriate activation function. The
uncertainty problem of ELM means that there exists an uncertainty
phenomenon between the small approximation error (or generalization
error) and high confidence of ELM estimator. As a result, it is
difficult to judge whether  a single time trail of ELM succeeds or
not. Concerning the generalization   degradation phenomenon, we find
that, with the widely used Gaussian-type activation function  (or
Gaussian kernel for the sake of brevity), ELM degrades the
generalization capability of FNN.

To facilitate the use of ELM, we   provide certain remedies to
circumvent the aforementioned drawbacks. On one hand, we find that
multiple times training can overcome  the  uncertainty problem of
ELM. On the  other hand, we show that, by adding   neurons and
implementing $l^2$ coefficient regularization  simultaneously,   the
generalization  degradation phenomenon of ELM can be avoided. In
particular, using $l^2$ coefficient regularization to determine the
connections with output neurons,
 ELM with Gaussian kernel can reach the almost optimal learning rate of
 FNN in the expectation sense, provided the regularization parameter is
appropriately tuned.

The study of this paper together with the conclusions in
\cite{Liu2013} provides a comprehensive feasibility analysis of ELM.
To be detail, the  performance of ELM depends heavily on the
activation function and the random assignment mechanism. With
appropriately selected activation function and random mechanism, ELM
does not degrade the generalization capability of FNN learning in
the expectation sense. However,  there also exist some activation
functions, with which  ELM degrades the generalization capability
for arbitrary random mechanism. Moreover, due to the randomness, ELM
suffers from an uncertainty problem, both in approximation and
learning. Our study also shows that both the uncertain problem and
degradation phenomenon are remediable. All these results lay a solid
fundamental for  ELM and give a guidance of how to use ELM more
efficiently.

The rest of the paper is organized as follows. After giving a fast
review of ELM, we present an uncertainty problem of ELM
approximation   in the next section. In Section 3, we first
introduce the main conception of statistical learning theory and
 then study the generalization capability of ELM with Gaussian kernel. We find that the deduced generalization error
bound is larger than that of FNN with Gaussian kernel. This means
that ELM with Gaussian kernel may degrade the generalization
capability. In Section 4, we provide a remedy to such a degradation.
Using the empirical covering number technique, we prove that
implementing $l_2$ coefficient regularization  can essentially
improve the generalization capability of ELM with Gaussian kernel.
In Section 5, we give proofs of the main results. We conclude the
paper in the final section with some useful remarks.

\section{An uncertainty problem of ELM approximation}

\subsection{Extreme learning machine}

The extreme learning machine (ELM), introduced by Huang et al.
\cite{Huang20061}   can be regarded as a two stage FNN  learning
system which  randomly assigns the connections with and within
hidden neurons in the first stage and tunes the connections with
output neurons in the second stage. Since then, various variants of
ELM such as evolutionary ELM \cite{Zhu2005}, Bayesian ELM
\cite{Soria2011}, incremental ELM \cite{Huang2007}, and regularized
ELM \cite{Deng2009}  were   proposed. We refer the readers to a
fruitful survey \cite{Huang20112}
 for more information about
ELM.

As a two stage learning scheme, ELM  comprises a choice of
hypothesis space, and a selection of optimization strategy (or
learning algorithm) in the first and second stages, respectively. To
be precise, in the first stage, ELM picks hidden parameters with and
within the hidden neurons randomly to build up the hypothesis space.
 This makes the hypothesis space
of ELM   form as
$$
        \mathcal H_{\phi,n}=\left\{\sum_{j=1}^na_j\phi(w_j,x): \ a_j\in\mathbf R\right\},
$$
where   $w_j$'s are drawn independently and identically (i.i.d.)
according to a specified distribution $\mu$. It is easy to see that
the hypothesis space of ELM is essentially a linear space. In the
second stage, ELM tunes the output weights by using the well
developed linear optimization technique. In this paper, we study the
generalization capability of the classical ELM \cite{Huang2006}
rather than its variants. That is, the linear optimization technique
employed in the second stage of ELM is the least square:
\begin{equation}\label{ELM Estimator}
          f_{{\bf z},\phi,n}=\arg\min_{f\in\mathcal
          H_{\phi,n}}\sum_{i=1}^m|f(x_i)-y_i|^2,
\end{equation}
where $(x_i,y_i)_{i=1}^m$ are the given samples.

\subsection{An uncertainty problem for ELM approximation}

The randomness of ELM  leads to a reduction of computational burden.
However, there also exists a certain defect caused by the
randomness. The main purpose of this section is to quantify such a
defect by studying the approximation capability of ELM with Gaussian
kernel.

For this purpose, we   introduce a quantity called the modulus of
smoothness \cite{DeVore1993} to measure the approximation
capability.
  The $r$-th modulus of smoothness
 \cite{DeVore1993} on $A\subseteq\mathbf R^d$ is  defined by
$$
                  \omega_{r,A}(f,t) = \sup_{|{\bf h}|_2\leq
                  t}\|\Delta_{{\bf h},A}^r(f,\cdot)\|_A,
$$
 where $|\cdot|_2$ denotes the Euclidean norm, $\|\cdot\|_A$
denotes the uniform norm on $C(A)$, and the $r$-th difference
$\Delta_{{\bf h},A}(f,\cdot)$ is defined by
$$
       \Delta_{{\bf h},A}^r(f,{ x})=\left\{
       \begin{array}{cc}
       \sum_{j=0}^r
       \left(^r_j\right)(-1)^{r-j}f({  x}+j{\bf h}) &
         \mbox{if}\ {  x}\in A_{r,{\bf h}}\\
         0 &
         \mbox{if}\ x\notin A_{r,{\bf h}}
         \end{array}\right.
$$
for ${\bf h}=(h_1,\dots,h_d)\in A^d$ and $A_{r,{\bf h}}:=\{{  x}\in
A:{ x}+s{\bf h}\in A,\ \mbox{for all}\ s\in[0,r]\}$. It is well
known \cite{DeVore1993} that
\begin{equation}\label{modulsproperty}
              \omega_{r,A}(f,t)\leq
              \left(1+\frac{t}{u}\right)^r\omega_{r,A}(f,u)
\end{equation}
for all $f\in C(A)$ and all $u>0$.

Let $s\in\mathbf N$, we focus on the following Gaussian-type
activation function (or Gaussian kernel),
\begin{equation}\label{Gaussian kernel}
         K_{\sigma,s}(t)=\sum_{j=1}^s\left(
         \begin{array}{c}
         s\\
         j
         \end{array}\right)(-1)^{1-j}\frac1{j^d }\left(\frac{2}{\sigma^2\pi}\right)^\frac{d}2\exp\left\{-\frac{2t^2}{j^2\sigma^2}\right\}.
\end{equation}
Then, the corresponding ELM estimator is defined by
\begin{equation}\label{Gaussian ELM Estimator}
          f_{{\bf z},\sigma,s,n}=\arg\min_{f\in\mathcal
          H_{\sigma,s,n}}\sum_{i=1}^m|f(x_i)-y_i|^2,
\end{equation}
where
$$
        \mathcal H_{\sigma,s,n}=\left\{\sum_{j=1}^na_jK_{\sigma,s}(\theta_j,x), \ x\in
        I^d\right\},
$$
$$
          K_{\sigma,s}(\theta_j,x):=K_{\sigma,s}((\theta_j-x)^2):=K_{\sigma,s}(|\theta_j-x|_2^2),
$$
 $I^d:=[0,1]^d$ and $\{\theta\}_{j=1}^n$ are
drawn i.i.d. according to arbitrary fixed distribution $\mu$ on the
interval $[-a,1+a]^d$ with $a>0$.

 The following  Theorem \ref{APPROXIMATION
ERROR} shows that there exists an uncertainty problem of ELM
approximation.

\begin{theorem}\label{APPROXIMATION ERROR}
 Let $d,s, n\in\mathbf N$. If $f\in C(I^d),$ then with
confidence at least $1-2\exp\{-cn\sigma^{2d}\}$ (with respect to
$\mu^n$), there holds
$$
      \inf_{g_n\in\mathcal H_{\sigma,s,n}}\|f-g_n\|_{I^d}\leq C
      \left(\omega_{s,  I^d}(f,\sigma)+
      \|f\|_{I^d}\sigma^d\right),
$$
where $C$ is a constant depending only on $d$ and $r$.
\end{theorem}

It follows from Theorem \ref{APPROXIMATION ERROR} that the
approximation capability of ELM with Gaussian kernel depends  on the
kernel parameters, $s$,  $\sigma$, and the number of hidden neurons,
$n$. Furthermore, Theorem \ref{APPROXIMATION ERROR} shows that,
compared with the classical FNN approximation, there exists an
additional uncertainty problem of ELM approximation. That is, both
the approximation error and the confidence  monotonously increase
with respect to $\sigma$. Therefore, it is impossible to deduce a
small approximation error with extremely high confidence. In other
words, it is difficult to judge whether the approximation error of
ELM is smaller than arbitrary specified approximation accuracy,
which does not appear in the classical Gaussian-FNN approximation
\cite{Xie2013}.

We find further in Theorem \ref{APPROXIMATION ERROR} that the best
choice of the kernel parameter, $\sigma$, is a trade-off between the
confidence and the approximation error. An advisable way
 to determine $\sigma$ is to set  $\sigma^{2d}=n^{\varepsilon-1}$ for
arbitrary small $\varepsilon\in\mathbf R_+$. Under this
circumstance, we can deduce that the approximation error of
$\mathcal H_{\sigma,s,n}$ asymptomatically equals to
$
           \omega_{s,I^d}(f,n^{(-1+\varepsilon)/(2d)})+n^{(-1+\varepsilon)/2}
$ with confidence at least $1-2\exp\{-cn^{\varepsilon}\}$. Finally,
we should   verify the optimality of the above approximation bound
and therefore justify the optimality of the selected $\sigma$. To
this end, we introduce the set of $r$th-smoothness functions.

Let     $u\in \mathbf{N}_{0}:=\{0\}\cup \mathbf{N}$, $v\in (0, 1]$,
and $r=u+v$. A function $f: I^d\rightarrow \mathbf{R}$ is said to be
$r$th-smooth if for every $\alpha =(\alpha
_{1},\cdots,\alpha_{d}),\alpha_{i}\in N_{0},$ $\sum
_{j=1}^{d}\alpha_{j}=u$, the partial derivatives $\frac{\partial
^{u}f}{\partial x{_{1}}^{\alpha_{1}}...\partial
x{_{d}}^{\alpha_{d}}}$ exist and satisfy
$$
             \left|\frac{\partial^{u}f}{\partial x{_{1}}^{\alpha_{1}}\cdots\partial
             x{_{d}}^{\alpha_{d}}}(x)
             -\frac{\partial ^{u}f}{\partial x{_{1}}^{\alpha_{1}}\cdots\partial
             x{_{d}}^{\alpha_{d}}}(z)\right |\leq c_0|x-z|^{v }_2,
$$
where $c_0$ is an absolute constant. Denote by $\mathcal{F}^r$ the
set of all $r$th-smooth functions. Furthermore, for arbitrary
$f\in\mathcal F^{r}$, it is easy to deduce \cite{DeVore1993} that
\begin{equation}\label{smooth estimation moduli}
           \omega_{s,  I^d}(f,t)\leq Ct^r,
\end{equation}
if $s\geq r$. According to Theorem \ref{APPROXIMATION ERROR} and
(\ref{smooth estimation moduli}), we obtain that
\begin{equation}\label{approximation error 11}
          \inf_{g_n\in\mathcal H_{\sigma,s,n}}\|f-g_n\|_{I^d}\leq
          Cn^\frac{-r+\varepsilon}{2d}
\end{equation}
holds with confidence at least $1-2\exp\{-cn^{\varepsilon}\}$ for
arbitrary $\varepsilon\in\mathbf R_+$, provided $f\in\mathcal F^r$,
$s\geq r$ and $r\leq d$.
 In the
following Proposition \ref{LOWER BOUND APPROXIMATION}, we show that
the approximation rate (\ref{approximation error 11}) can not be
essentially improved, at least for the univariate case.

\begin{proposition}\label{LOWER BOUND APPROXIMATION}
Let $d=s=1,n\in\mathbf N$, $\beta>0$, $0<\varepsilon<1$ and
$r=1-\varepsilon$. If
 $f_\rho\in\mathcal F^{r}$ and $\sigma= n^{(-1+\varepsilon)/2}$,
 then with confidence at least $1-2\exp\{-cn^{\varepsilon}\}$ (with respect to $\mu^n$), there
 holds
\begin{equation}\label{Optimality}
      C_1 n^{-\frac{r}{2}-\varepsilon}\leq\sup_{f\in\mathcal F^r}\inf_{g_n\in\mathcal
       H_{\sigma,r,n}}\|f-g_n\|_{I^d}\leq C_2 n^\frac{-r+\varepsilon}{2}.
\end{equation}
\end{proposition}

\section{A generalization   degradation problem of ELM with Gaussian kernel}
Along the  flavor of  \cite{Liu2013}, we also   analyze   the
feasibility of ELM in the framework of statistical learning theory
\cite{Cucker2001}. We find in this section that there exists a
generalization degradation phenomenon of ELM. In particular, unlike
\cite{Liu2013}, the result in this section shows that ELM with
Gaussian kernel degrades the generalization capability of FNN.

\subsection{A fast review of statistical learning theory}

 Let $M> 0$,
$X=I^d$, $Y\subseteq [-M,M]$ be the input and output spaces,
respectively. Suppose that  ${\bf z}=(x_i,y_i)_{i=1}^m$ is a finite
set of  random samples drawing   i.i.d. according to an unknown but
definite distribution
 $\rho$, where
 $\rho$ is assumed to admit
the decomposition
$$
                    \rho(x,y)=\rho_X(x)\rho(y|x).
$$
Suppose further that $f:X\rightarrow Y$ is a function that one uses
to model the correspondence between $X$ and $Y$, as induced by
$\rho$. One natural measurement of the error incurred by using $f$
of this purpose is the generalization error, defined by
$$
                     \mathcal E(f):=\int_Z(f(x)-y)^2d\rho,
$$
which is minimized by the regression function \cite{Cucker2001},
defined by
$$
                     f_\rho(x):=\int_Yyd\rho(y|x).
$$
We do not know this ideal minimizer $f_\rho$, since $\rho$ is
unknown, but we have access to random examples from $X\times Y$
sampled according to $\rho$. Let $L^2_{\rho_{_X}}$ be the Hilbert
space of $\rho_X$ square integrable function on $X$, with norm
denoted by $\|\cdot\|_\rho.$  Then for arbitrary $f\in
L^2_{\rho_{_X}},$ there holds
\begin{equation}\label{equality}
                     \mathcal E(f)-\mathcal E(f_\rho)=\|f-f_\rho\|^2_\rho
\end{equation}
 with the assumption   $f_\rho\in
L^2_{\rho_{_X}}$.

\subsection{The generalization capability of ELM with Gaussian
kernel}

Let $\pi_Mf(x)=\min\{M,|f(x)|\}\mbox{sgn}(f(x))$ be the truncation
operator on $f(x)$ at level $M$.
 As $y\in [-M,M]$, it is easy to check \cite{Zhou2006} that
$$
          \|\pi_M f_{{\bf z},\sigma,s,n}-f_\rho\|_\rho\leq\|f_{{\bf
          z},\sigma,s,n}-f_\rho\|_\rho.
$$
 Thus, the aim of this  section is to
bound
\begin{equation}\label{target}
                     \mathcal E(\pi_M f_{{\bf
                     z},\sigma,s,n})-\mathcal E(f_\rho)=\|\pi_M f_{{\bf
                     z},\sigma,s,n}-f_\rho\|_\rho^2.
\end{equation}
The error (\ref{target}) clearly depends on ${\bf z}$ and therefore
has a stochastic nature. As a result, it is  impossible to say
anything about (\ref{target}) in general for a fixed ${\bf z}$.
Instead, we can look at its behavior in statistics as measured by
the expected error
$$
               \mathbf E_{\rho^m}(\|\pi_M f_{{\bf
                     z},\sigma,s,n}-f_\rho\|_\rho):=\int_{Z^m}\|\pi_M f_{{\bf
                     z},\sigma,s,n}-f_\rho\|d\rho^m,
$$
where the expectation is taken over all realizations ${\bf z}$
obtained for a fixed $m$, and $\rho^m$ is the $m$ fold tensor
product of $\rho$. In following Theorem \ref{THEOREM LEARNG}, we
give  an upper bound estimate for  (\ref{target}) in the sense of
expectation.

\begin{theorem}\label{THEOREM LEARNG}
Let $d,s,n,m\in\mathbf N$, $\varepsilon>0$, $r\in\mathbf R$ and
$f_{{\bf z},\sigma,s,n}$ be defined as in (\ref{Gaussian ELM
Estimator}). If $f_\rho\in\mathcal F^{r}$  with $r\leq s$,
 $\sigma=m^{ \frac{-1+\varepsilon}{2r+2d}}$ and
 $n=\left[m^{\frac{d}{r+d}}\right]$,
 then with
probability at least $1-2\exp\{-cm^{\frac{\varepsilon d}{d+r}}\}$
(with respect to $\mu^n$), there holds
\begin{equation}\label{Learning rate}
        \mathbf E_{\rho^m}(\|\pi_Mf_{{\bf z},\sigma,s,n}-f_\rho\|_\rho^2)\leq
        C\left(m^{-\frac{(1-\varepsilon)r}{r+d}}\log m
          +m^{-\frac{d(1-\varepsilon)}{r+d}}\right),
\end{equation}
where $[t]$ denotes the  integer part of the real number $t$,  $c$
and $C$ are constants depending only on $M$, $s$, $r$ and $d$.
\end{theorem}

It can be found in Theorem \ref{THEOREM LEARNG} that  a new quantity
$\varepsilon$ is introduced to quantify the randomness of ELM. It
follows from (\ref{Learning rate}) that $\varepsilon$ describes the
uncertainty between the confidence and generalization capability.
That is, we cannot obtain both extremely small generalization error
and high confidence. This means that there also exists an
uncertainty problem for ELM learning. Accordingly,  Theorem
\ref{THEOREM LEARNG}  shows that it is reasonable  to choose a very
small $\varepsilon$, under which circumstance, we can deduce a
learning rate close to $m^{-\frac{r}{r+d}}\log m$ with a tolerable
confidence, provided $r\leq d$.

Before drawing the conclusion that ELM with Gaussian kernel degrades
the generalization capability, we should verify the optimality of
both the established learning rate (\ref{Learning rate})   and the
selected parameters such as $\sigma$ and $n$. We begin the analysis
by illustrating the optimality of the learning rate deduced in
(\ref{Learning rate}). For this purpose, we give the following
Proposition \ref{LOWER BOUND LEARNING}.

\begin{proposition}\label{LOWER BOUND LEARNING}
Let $d=s=1,n,m\in\mathbf N$, $\beta>0$, $0<\varepsilon<1$,
$r=1-\varepsilon$ and $f_{{\bf z},\sigma,s,n}$ be defined as in
(\ref{Gaussian ELM Estimator}). If
 $f_\rho\in\mathcal F^{r}$, $\sigma=m^{ \frac{-1+\varepsilon}{2r+2}}$ and
$n=\left[m^{\frac{1}{1+r}}\right]$, then with probability at least
$1-2\exp\{-cm^{\frac{\varepsilon  }{1+r}}\}$ (with respect to
$\mu^n$), there holds
\begin{equation}\label{Optimality}
       C_1m^{-\frac{r}{1+r}}\leq \mathbf E(\|\pi_Mf_{{\bf z},\sigma,s,n}-f_\rho\|_\rho^2)\leq
        C_2m^{-\frac{r(1-\varepsilon)}{1+r}}\log m,
\end{equation}
where $c$, $C_1$ and $C_2$ are constants depending only on $r$ and
$M$.
\end{proposition}

Modulo an arbitrary small number  $\varepsilon$ and the logarithmic
factor, the upper and lower bounds of (\ref{Optimality}) are
asymptomatically identical. Therefore, the established learning rate
in Theorem \ref{THEOREM LEARNG} is almost essential. This means that
the established learning rate (\ref{Learning rate}) can not be
essentially improved, at least for the univariate case.

Now, we turn to justify the optimality of the selections of $\sigma$
and $n$ in Theorem \ref{THEOREM LEARNG}. The optimality of $\sigma$
can be directly derived from the uncertainty problem of ELM. To be
detail, according to Theorem \ref{APPROXIMATION ERROR} and
Proposition \ref{LOWER BOUND APPROXIMATION}, the optimal selection
of $\sigma$ is to set $\sigma=n^\frac{\varepsilon-1}{2d}$. Noting
that $n=\left[m^\frac{d}{d+r}\right]$, it is easy to deduce that the
optimal selection of $\sigma$ is $m^\frac{-1+\varepsilon}{2r+2d}$.
Finally, we show the optimality of the parameter $n$. The main
principle  to qualify it is the known ``bias and variance`` dilemma
\cite{Cucker2001}, which declares     that a small $n$ may derive a
large bias (approximation error), while a large $n$   deduces a
large variance (sample error).  The best $n$ is thus obtained when
the best comprise between the conflicting requirements of small bias
and small variance is achieved.  In the proof of Theorem
\ref{THEOREM LEARNG}, we can find that  the quantity
$n=\left[m^{d/(r+d)}\right]$ is selected to balance the
approximation and sample errors. Therefore, we can conclude that $n$
is optimal in the sense of ``bias and variance'' balance.

Based on the above assertions, we compare Theorem \ref{THEOREM
LEARNG} with some related work and  propose  then the main viewpoint
of this section. Imposing the same smooth assumption on the
regression function, the optimal learning rate of the FNN with
Gaussian kernel
  was established in \cite{Lin2013b}, where  Lin et
al. deduced that FNNs can achieve the learning rate as
$m^{-2r/(2r+d)}\log m$. They also showed that there are
  $\left[m^{d/(2r+d)}\right]$ neurons needed to deduce the almost
optimal learning rate. Similarly, Eberts and Steinwart
\cite{Eberts2011}
 have also built an almost optimal learning rate analysis for the support vector machine (SVM) with the Gaussian
kernel. They showed that, modulo an arbitrary small number, both the
upper and lower bounds of learning rate of SVM with Gaussian can
also attain the optimal learning rate, $m^{-2r/(2r+d)}$.  However,
Theorem \ref{THEOREM LEARNG} and Proposition \ref{LOWER BOUND
LEARNING} imply that the learning rate of  ELM with Gaussian kernel
can not be faster than $m^{-r/(r+d)}$. Noting
$m^{-2r/(2r+d)}<m^{-r/(r+d)}$ and $m^{d/(2r+d)}<m^{d/(r+d)}$, we
find that the prediction accuracy of ELM with Gaussian kernel is
much larger than that of FNN even though
  more neurons are used in ELM. Furthermore, it should be pointed
  out that if the numbers of utilized neurons in ELM and FNN are
  identical, then the learning rate of ELM is even worse. Indeed, if
$n=\left[m^{d/(2r+d)}\right]$, then the learning rate of ELM with
Gaussian kernel can not be faster than $m^{-r/(2r+d)}$\footnote{The
proof of this conclusion is the same as that of Theorem \ref{THEOREM
LEARNG}, we omit it for the sake of brevity.}.
  Therefore, we can draw the conclusion that
 ELM with
Gaussian kernel degrades the generalization capability.

\section{Remedy of the degradation}
As is shown in the previous section,    ELM with inappropriately
selected activation function suffers from the uncertainty problem
and generalization degradation phenomenon. To circumvent the former
one, we can employ a multiple training strategy which has already
been proposed in \cite{Liu2013}. The main focus of this section is
to tackle the generalization capability degradation phenomenon.
 For
this purpose, we use  the $l^2$ coefficient regularization strategy
\cite{Wu2008} in the second stage of ELM. That is, we implement the
following strategy to build up the ELM estimator:
\begin{equation}\label{regularization}
           f_{{\bf z},\sigma,s,\lambda,n}=\arg\min_{f\in\mathcal H_{\sigma,s,n}}\left\{\frac1m\sum_{i=1}^m(f(x_i)-y_i)^2+\lambda\Omega(f)\right\},
\end{equation}
where $\lambda=\lambda(m)>0$ is a regularization parameter and
$$
          \Omega(f)= \sum_{i=1}^m|a_i|^2, \ \mbox{for}\ f=\sum_{i=1}^na_iK_{\sigma,s}(\theta_i,x)\in \mathcal H_{\sigma,s,n}.
$$

The following theorem shows that the generalization capability of
ELM with Gaussian kernel can be essentially improved by using the
regularization technique, provided the number of neurons is
appropriately adjusted.

\begin{theorem}\label{REGULARIZATION THEOREM}
Let  $d,s,n,m\in\mathbf N$, $\varepsilon>0$ and  $f_{{\bf
z},\sigma,s,\lambda,n}$ be defined in     (\ref{regularization}). If
$f_\rho\in\mathcal F^r$ with $d/2\leq r\leq d$,
$\sigma=m^{-\frac1{2r+d}+\varepsilon}$,
$n=\left[m^\frac{2d}{2r+d}\right]$, $s\geq r$ and
$\lambda=m^{-\frac{2r-d}{4r+2d}}$, then with confidence at least
$1-2\exp\{-cm^{\frac{\varepsilon d}{d+r}}\}$ (with respect to
$\mu^n$), there holds

\begin{equation}\label{th2}
           C_1m^\frac{-2r}{2r+d}\leq \mathbf E_{\rho^m}\|\pi_Mf_{{\bf z},\sigma,s,\lambda,n}-f_\rho\|_\rho^2\leq
           C_2m^{-\frac{2r}{2r+d}+\varepsilon}\log m,
\end{equation}
 where $C_1$ and $C_2$ are   constants depending
only on $d$, $r$, $s$ and $M$.
\end{theorem}

Theorem \ref{REGULARIZATION THEOREM} shows that, up to an arbitrary
small real number $\varepsilon$ and the logarithmic factor, the
regularized  ELM estimator (\ref{regularization}) can achieve a
learning rate as fast as $m^{-2r/(2r+d)}$ with high probability.
Noting that $m^{-2r/(2r+d)}<m^{-r/(r+d)}$ we can draw the conclusion
that $l^2$ coefficient regularization technique can essentially
improve the generalization capability of ELM with Gaussian kernel.
Furthermore, as is shown above, the best learning rates of both SVM
and FNN with Gaussian kernel   asymptomatically equal to
$m^{-2r/(2r+d)}$. Thus, Theorem \ref{REGULARIZATION THEOREM}
illustrates that the regularization technique not only improves the
generalization capability of ELM with Gaussian kernel, but also
optimizes its generalization capability. In other words,
implementing $l^2$ coefficient regularization   in the second stage,
ELM with Gaussian kernel can be regarded as an almost optimal FNN
learning strategy.

However, it should also be pointed out that the utilized neurons of
regularized ELM is much larger than that of the FNN. Indeed, to
obtain the same optimal learning rate, $m^{-2r/(2r+d)}$, there are
$\left[m^{2d/(2r+d)}\right]$ neurons required in ELM with Gaussian
kernel, while the number of utilized neurons in the traditional FNN
learning is $\left[m^{d/(2r+d)}\right]$.  Therefore, although
regularized ELM can attain the almost optimal learning rate with
high probability, the price to obtain such a rate is  higher than
that of FNN.

\section{Proofs}
\subsection{Proof of Theorem \ref{APPROXIMATION ERROR}}
To prove Theorem \ref{APPROXIMATION ERROR}, we need the following
nine lemmas. The first one can be found in \cite{Lin2013a}, which is
an extension of Lemma 2.1 in \cite{Xie2013}.

\begin{lemma}\label{TOPIC}
 Let $f\in C({I^d})$. There exists an $F\in C(\mathbf{R}^d)$
satisfying
$$
                               F({  x})=f({  x}),\ {  x}\in I^d
$$
such that for arbitrary ${  x}\in I^d, \|{\bf h}\|<\delta\leq 1$,
there holds
$$
                    \|F\|_\infty:=\sup_{{ x}\in \mathbf {R}^d}|F({  x})|\leq \|f\|=\sup_{{  x}\in I^d}|f({  x})|
$$
and
\begin{equation}\label{relation111}
                     \omega_{r,\mathbf R^d}(F,\delta)   \leq \omega_{r,I^d}(f,\delta).
\end{equation}
\end{lemma}

To state the next lemma, we should introduce a convolution operator
concerning the kernel $K_{\sigma,s}$. Denote
$$
               K_{\sigma,s}\ast F({  x}):=\int_{\mathbf R^d}F({  y})K_{\sigma,s}({  x}-{
               y})d{ y}.
$$
The following Lemma 2 gives an error estimate for the deviation of
continuous function and its Gaussian convolution, which can be
deduced from \cite[Theorem 2.2]{Eberts2011}.

\begin{lemma}\label{OPEATOR}
 Let $F\in C(\mathbf R^d)$ be a bounded and uniformly
continuous function defined on $\mathbf R^d$. Then,
\begin{equation}\label{convolution}
                          \|F-K_{\sigma,s}\ast F\|_\infty \leq C_s \omega_{s,\mathbf R^d}(F,\sigma).
\end{equation}
\end{lemma}

Let $J$ be arbitrary compact subset of $\mathbf R^d$. For $l\geq0$,
denote by $\mathcal T_l^d$ the set of trigonometric polynomials
defined on $J$ with degree at most $l$. The following Nikol'skii
inequality can be found in \cite{Borwein1995}.

\begin{lemma}\label{NIKOLSKII}
 Let $1\leq p<q\leq\infty,$ $l\geq1$ be an integer, and
$T_l\in\mathcal T_l^d$. Then
$$
          \|T_l\|_{L^q(J)}\leq
          Cl^{\frac{d}{p}-\frac{d}{q}}\|T_l\|_{L^p(J)},
$$
where the constant $C$ depends only on $d$.
\end{lemma}

For further use, we also should introduce the following
probabilistic  Bernstein inequality for random variables, which can
be found in \cite{Cucker2001}.

\begin{lemma}\label{BERNSTEIN}
 Let $\xi$ be a random variable on a probability space $Z$ with
mean $E(\xi)$, variance $\gamma^2(\xi)=\gamma^2_\xi$. If
 $|\xi(z)-E(\xi)|\leq M_\xi$ for almost all ${\bf z}\in Z$. then, for all $\varepsilon>0$,
$$
           \mathbf P\left\{\left|\frac1n\sum_{i=1}^n\xi(z_i)-E(\xi)\right|
           \geq\varepsilon\right\}\leq 2\mbox{exp}\left\{-\frac{n\varepsilon^2}{2\left(\gamma^2_\xi+\frac13M_\xi\varepsilon\right)}\right\}.
$$
\end{lemma}

By the help of Lemma \ref{NIKOLSKII} and Lemma \ref{BERNSTEIN}, we
are in a position to give the following probabilistic
Marcinkiewicz-Zygmund  inequality  for trigonometric polynomials.

\begin{lemma}\label{MZTRIGNOMETRIC}
Let $J$ be a compact subset of $\mathbf R^d$ and $0<p\leq\infty$. If
$\Xi=\{\theta_i\}_{i=1}^n$ is a set of i.i.d. random variables drawn
on $J$  according to arbitrary distribution $\mu$ , then
\begin{equation}\label{MZ}
           \frac12\|T_l\|_p^p\leq\frac1n\sum_{i=1}^n|T_l(\theta_i)|^p\leq\frac{3}2\|T_l\|_p^p,\quad
           \forall T_l\in\mathcal T_l^{d}
\end{equation}
holds with probability at least
$$
           1-2\mbox{exp}\left\{-\frac{C_pn}{l^d}\right\},
$$
where $C_p$ is a constant depending only on $d$ and $p$.
\end{lemma}

\begin{IEEEproof}  Since we model the sampling set $\Xi$ is a sequence of i.i.d.
random variables in $J$, the sampling points are a sequence of
functions $\theta_j=\theta_j(\omega)$ on some probability space
$(\Omega,\mathbf P)$. Without loss of generality, we assume
$\|T_l\|_p=1$ for arbitrary fixed $p$. If we set
$\xi_j^p(T_l)=|T_l(\theta_j)|^p$, then we have
$$
           \frac1n\sum_{i=1}^n|T_l(\theta_i)|^p-
             E\xi_j^p=\frac1n\sum_{i=1}^n|T_l(\theta_i)|^p-\|T_l\|_p^p,
$$
where we use the equality
$$
            E\xi_j^p=\int_{\Omega}|T_l(\eta(\omega_j))|^pd\omega_j=\int_{J}|T_l(\theta)|^pd\theta=\|T_l\|_p^p=1.
$$
Furthermore,
$$
            |\xi_j^p-E\xi_j^p|\leq\sup_{\omega\in\Omega}\left||T_l(\theta(\omega))|^p-\|T_l\|_p^p\right|
            \leq \|T_l\|_\infty^p-\|T_l\|_p^p.
$$
It follows from Lemma \ref{NIKOLSKII} that
$$
               \|T_l\|_\infty\leq Cl^{\frac{d}{p}}\|T_l\|_p=Cl^{\frac{d}{p}}.
$$
Hence
$$
             |\xi_j^p-E\xi_j^p|\leq (Cl^d-1).
$$
On the other hand, we have
\begin{eqnarray*}
              \gamma_\xi^2
              &=&
              E((\xi_j^p)^2)-(E(\xi_j^p))^2\\
              &=&
              \int_\Omega|T_l(\theta(\omega))|^{2p}d\omega-\left(\int_\Omega|T_l(\theta(\omega))|^pd\omega\right)^2\\
              &=&
              \|T_l\|_{2p}^{2p}-\|T_l\|_p^{2p}.
\end{eqnarray*}
Then using Lemma \ref{NIKOLSKII} again, there holds
$$
            \gamma^2_\xi\leq
            Cl^{2dp(\frac1p-\frac1{2p})}\|T_l\|_p^{2p}-\|T_l\|_p^{2p}=(Cl^d-1).
$$
Thus it follows from Lemma \ref{BERNSTEIN} that with confidence at
least
\begin{eqnarray*}
              &&1-2\mbox{exp}\left\{-\frac{n\varepsilon^2}{2\left(\gamma^2+\frac13M_\xi\varepsilon\right)}\right\}\\
              &\geq&
              1-2\mbox{exp}\left\{-\frac{n\varepsilon^2}{2\left((Cl^d-1)
              +\frac13(
              Cl^d-1)\varepsilon\right)}\right\},
\end{eqnarray*}
 there holds
$$
              \left|\frac1n\sum_{i=1}^n|T_l(\theta_i)|^p-\|T_l\|_p^p\right|\leq\varepsilon.
$$
This means that if $X$ is a sequence of i.i.d. random variables,
then the Marcinkiewicz-Zygmund inequality
$$
           (1-\varepsilon)\|T_l\|_p^p\leq\frac1n\sum_{i=1}^n|T_l(\theta_i)|^p\leq(1+\varepsilon)\|T_l\|_p^p\quad
$$
holds with probability at least
$$
           1-2\mbox{exp}\left\{-\frac{cn\varepsilon^2}{l^d(1+\varepsilon)}\right\}.
$$
Then (\ref{MZ}) is verified by setting $\varepsilon=\frac12$.
\end{IEEEproof}

To state the next lemma, we need introduce the following
definitions. Let $\mathcal X$ be a finite dimensional vector space
with norm $\|\cdot\|_{\mathcal X}$, and $\mathcal Z\subset \mathcal
X^*$ be a finite set. We say that $\mathcal Z$ is a norm generating
set for $\mathcal X$ if the mapping $T_{\mathcal Z}: \mathcal
X\rightarrow\mathbf R^{Card(\mathcal Z)}$ defined by $T_{\mathcal
Z}(x)=(z(x))_{z\in \mathcal Z}$ is injective, where $Card(\mathcal
Z)$ is the cardinality of the set $\mathcal Z$ and $T_{\mathcal Z}$
is named as the sampling operator. Let $W:=T_{\mathcal Z}(\mathcal
X)$ be the range of $T_{\mathcal Z}$, then the
 injectivity of $T_{\mathcal Z}$ implies that $T_{\mathcal Z}^{-1}:W\rightarrow \mathcal X$ exists.
 Let $\mathbf R^{Card(\mathcal Z)}$ have a norm $\|\cdot\|_{\mathbf R^{Card(\mathcal Z)}}$, with
 $\|\cdot\|_{\mathbf R^{Card(\mathcal Z)^*}}$ being its dual norm on $\mathbf
 R^{Card(\mathcal Z)^*}$. Equipping $W$ with the induced norm, and let
 $\|T_{\mathcal Z}^{-1}\|:=\|T_{\mathcal Z}^{-1}\|_{W\rightarrow \mathcal
 X}.$  In addition, let
 $\mathcal K_+$ be the positive cone of $\mathbf R^{Card(\mathcal Z)}$: that is, all
 $(r_z)\in\mathbf R^{Card(\mathcal Z)}$ for which $r_z\geq0$. Then
 the following Lemma \ref{NORMING SET} can be found in  \cite{Mhaskar2000}.

\begin{lemma}\label{NORMING SET}
  Let $\mathcal Z$ be a norm generating set for $\mathcal X$,
with $T_{\mathcal Z}$ being the corresponding sampling operator. If
$y\in \mathcal X^*$ with $\|y\|_{\mathcal X^*}\leq A$, then there
exist real numbers $\{a_z\}_{z\in \mathcal Z}$, depending only on
$y$ such that for every $x\in  \mathcal X,$
$$
                       y(x)=\sum_{z\in \mathcal Z}a_zz(x),
$$
and
$$
               \|(a_z)\|_{\mathbf R^{Card(\mathcal Z)^*}}\leq A\|T_{\mathcal Z}^{-1}\|.
$$
Also, if $W$ contains an interior point $v_0\in \mathcal K_+$ and if
$y(T_{\mathcal Z}^{-1}v)\geq0$ when $v\in V\cap \mathcal K_+$, then
we may choose $a_z\geq 0.$
\end{lemma}

Using Lemma \ref{NORMING SET} and Lemma \ref{MZTRIGNOMETRIC}, we can
deduce the following probabilistic numerical integral formula for
trigonometric polynomials.

\begin{lemma}\label{CUBATURE TRI}
 Let $J$ be a compact subset of $\mathbf R^d$. If
$\Xi=\{\theta_i\}_{i=1}^n$ are i.i.d. random variables drawn
according to arbitrary distribution $\mu$, then there exists a set
of real numbers $\{c_i\}_{i=1}^n$ such that
$$
              \int_JT_l({ x})d{  x}=\sum_{i=1}^nc_iT_l(\theta_i),\
              \forall T_l\in\mathcal T_l^d
$$
holds with confidence at least
$$
           1-2\mbox{exp}\left\{-\frac{C_1n}{l^d}\right\},
$$
subject to
$$
                \sum_{i=1}^n|c_i|^2\leq C/n,
$$
where $C_1$   and $C$ are constants depending only on $d$.
\end{lemma}

\begin{IEEEproof}   In Lemma \ref{NORMING SET}, we
take $\mathcal X=\mathcal T_l^d$, $\|T_l\|_{\mathcal X}=\|T_l\|_p$,
and $\mathcal Z$ to be the set of point evaluation functionals
$\{\delta_{\theta_i}\}_{i=1}^n$. The operator $T_{\mathcal Z}$ is
then the restriction map $T_l\mapsto T_l|_{\Xi},$ with
$$
            \|f\|_{\Xi,p}^p:=\left\{\begin{array}{cc}
             \left(\frac1n\sum_{i=1}^n|f(\theta_i)|^p\right)^\frac1p, &
             0<p<\infty,\\
            \sup_{1\leq i\leq n}\{|f(\theta_i)|\}, & p=\infty.
            \end{array}
            \right.
$$
  It follows from
Lemma \ref{MZTRIGNOMETRIC} with $p=2$ that with confidence at least
$$
           1-2\mbox{exp}\left\{-\frac{Cn}{l^d}\right\}
$$
there holds $\|T_{\mathcal Z}^{-1}\|\leq2. $ We now take $y$ to be
the functional
$$
              y: T_l\mapsto \int_JT_l({ x})d{  x}.
$$
By H\"{o}lder inequality, $\|y\|_{\mathcal X^*}\leq |J|$, where
$|J|$ denotes the volume of $J$. Therefore, Lemma \ref{NORMING SET}
shows that
$$
              \int_I T_l({ x})d{
              x}=\sum_{i=1}^nc_iT_l(\theta_i)
$$
holds with confidence at least
$$
           1-2\mbox{exp}\left\{-\frac{C_pn}{l^d}\right\},
$$
subject to
$$
                \frac1n\sum_{i=1}^n\left(\frac{|c_i|}{1/n}\right)^2\leq 2{|J|}.
$$
Therefore, we obtain that $\sum_{i=1}^n|c_i|^2\leq C/n$, where $C$
is a constant depending only on $d$.
\end{IEEEproof}

 Let $B=[-a,1+a]^d$ and $\mathcal P_l^d$ be the class of algebraic
 polynomials defined on $B$ with degree at most $l$. By the help of
the above lemma, we can get the following probabilistic numerical
integral formula for algebraic polynomials.

\begin{lemma}\label{CUBATURE ALG}
   If
$\Xi=\{\eta_i\}_{i=1}^n$ are i.i.d. random variables drawn according
to arbitrary distribution $\mu$, then there exists a set of real
numbers $\{a_i\}_{i=1}^n$ such that
$$
              \int_BP_l({  x})d{  x}=\sum_{i=1}^na_iP_l(\eta_i),\
              \forall  P_l\in\mathcal P_l^d
$$
holds with confidence at least
$$
           1-2\mbox{exp}\left\{-\frac{C_1n}{l^d}\right\},
$$
subject to
$$
                \sum_{i=1}^m|a_i|^2\leq \frac{C}{n},
$$
where $C_1$ and $C$ are  constanst depending only on  $d$.
\end{lemma}

\begin{IEEEproof} Since $x=(x_{(1)},\dots,x_{(d)})$, we have
$$
       \int_Bf(x)dx=\int_{-a}^{1+a}\cdots\int_{-a}^{1+a}f(x_{(1)},\dots,x_{(d)})dx_{(1)}\cdots
       dx_{(d)}.
$$
Set $x_{(i)}=(1+|a|)\cos v_i$, $i=1,\dots,d$, then we have
\begin{eqnarray*}
           &&\int_BP_l(x)dx
           =
           \int_{-a}^{1+a}\cdots\int_{-a}^{1+a}P_l((1+|a|)\\
           &\times&
           \cos v_1,\dots,(1+|a|)\cos v_d)\\
           &\times&
           d(1+|a|)\cos
           v_1\cdots
                 d(1+|a|)\cos v_d
           =
           \int_{J_a} T_{l+d}(v)dv,
\end{eqnarray*}
where $J_a$ is a compact subset of $\mathbf R^d$ and
\begin{eqnarray*}
        T_{l+d}(v)
        &=&
        (-(1+|a|))^dP_l((1+|a|)\cos v_1,\dots,(1+|a|)\\
        &\times&
        \cos
        v_d)\sin v_1\cdots\sin v_d.
\end{eqnarray*}
Hence, $ T_{l+d}\in \mathcal T_{l+d}^d$ and then Lemma \ref{CUBATURE
ALG} can  be directly deduced from   Lemma \ref{CUBATURE TRI}.
\end{IEEEproof}

By using   Lemma \ref{CUBATURE ALG}, we can deduce the following
error estimator.

\begin{lemma}\label{ESTIMATOR1}
 Let $a>0$, $u,l\in\mathbf N$. If $\Xi:=\{\eta_i\}_{i=1}^n$ is
a random variable drawing identically and independently according to
$\mu$ on $[-a,1+a]$, then with confidence at least
$1-2\exp\{-cn/(u+l)^d\},$ there holds
\begin{eqnarray*}
      &&\inf_{g_n\in\mathcal H_{\sigma,s,n}}\|K_{\sigma,s}\ast F-g_n\|\\
      &\leq&
       C_r \left(\omega_{s,I^d}(f,1/l)+
      a\|f\|\sigma^d
      +
   \sigma^{-d} \frac{2^u}{ u!\sigma^2}\right),
\end{eqnarray*}
where $C_s$ is a constant depending only on $d$ and $s$.
\end{lemma}

\begin{IEEEproof} For arbitrary
$f\in C(I^d)$, let $F$ and $K_{\sigma,s}\ast F$ defined as in Lemma
\ref{TOPIC} and Lemma \ref{OPEATOR}, respectively. Then,
\begin{eqnarray*}
            &&K_{\sigma,s}\ast F
            =
            \int_{\mathbf R^d}K_{\sigma,s}({  x}-{ y})F({  y})
            d{  y}\\
           & =&
            \int_BK_{\sigma,s}({ x}-{  y})F({  y})
            d{  y}
            +
            \int_{\mathbf R^d-B}K_{\sigma,s}({  x}-{  y})F({  y})
            d{  y}.
\end{eqnarray*}
At first, we give an upper bound estimate for $\int_{\mathbf
R^d-B}K_{\sigma,s}({  x}-{  y})F({  y})
            d{  y}$.
It follows from Lemma \ref{TOPIC} and the definition of
$K_{\sigma,s}$ that
\begin{eqnarray*}
         &&\left|\int_{\mathbf R^d-B}K_{\sigma,s}({ x}-{ y})F({  y})
            d{  y}\right|\\
            &\leq&
             \|f\|_{I^d}\sum_{j=1}^s\left(\begin{array}{c} s\\
             j\end{array}\right)\frac1{j^d}\left(\frac{2}{\sigma^2\pi}\right)^{d/2}\\
             &\times&
             \int_{\mathbf{R}^d-B}
               \exp\left\{-\frac{2\|{  x}-{ y}\|_2^2}{j^2\sigma^2}\right\}d{ y}\\
                &\leq&
                \|f\|_{I^d}\sum_{j=1}^s\left(\begin{array}{c} s\\
             j\end{array}\right)\frac1{j^d}\left(\frac{2}{\sigma^2\pi}\right)^{d/2}\\
             &\times&
             \left(\left(\int_{-\infty}^{-a}+\int_a^\infty\right)
               \exp\left\{-\frac{2t^2}{j^2\sigma^2}\right\}dt\right)^d\\
               &\leq&
                2\|f\|_{I^d}\sum_{j=1}^s\left(\begin{array}{c} s\\
             j\end{array}\right)\frac1{j^d}\left(\frac{2}{\sigma^2\pi}\right)^{d/2}\\
             &\times&
             \left(\int_a^\infty
               \exp\left\{-\frac{2at}{j^2\sigma^2}\right\}dt\right)^d\\
               &\leq&
               C_s \|f\|_{I^d}a^{-1}\sigma^d,
\end{eqnarray*}
where $C_s$ is a constant depending only on $d$ and $r$.

On the other hand, for $F\in C(B)$ and $s\in \mathbf N$, it is well
known \cite{DeVore1993} that there exists a $P_l\in \mathcal{P}_l^d$
and absolute constants $C_1, C_2$ such that
\begin{equation}\label{Jackson}
                        \|F-P_l\|\leq C_1 \inf_{P\in \mathcal{P}^d_l}\|F-P\|_B=:C_1E_l(F),
\end{equation}
and
\begin{equation}\label{Jackson1}
 \|P_l\|_B\leq
                             C_2\|F\|_B\leq C_2\|f\|_{I^d}.
\end{equation}
  Then, for arbitrary
$\{b_i\}_{i=1}^{n}\subset \mathbf R$, there holds
\begin{eqnarray}\label{decomposition}
               &&\int_{B}F({  y})K_{\sigma,s}({  x}-{  y})d{  y}-\sum_{i=1}^nb_iK_{\sigma,s}({  x}-\eta_i)
               \nonumber\\
             &=&\int_{B}(F({  y})-P_l({  y}))K_{\sigma,s}({  x}-{
             y})d{  y}\nonumber\\
             &+&
             \int_{B}P_l({ y})K_{\sigma,s}({  x}-{  y})d{  y}-\sum_{i=1}^nb_iK_{\sigma,s}({  x}-\eta_i).
\end{eqnarray}
Let $u\in\mathbf N$. Then, for arbitrary univariate algebraic
polynomial $q$ of degree not larger than $u$, we obtain
\begin{eqnarray*}
               &&\int_{B}P_l({  y})K_{\sigma,s}({  x}-{  y})d{  y}-\sum_{i=1}^nb_iK_{\sigma,s}({  x}-\eta_i)\\
               &=&
               \int_{B}P_l({  y})
               (K_{\sigma,s}({  x}-{  y})-q({  x}-{  y}))dt \\
               &+&
               \int_{B}P_l(y)q({  x}-{  y})dy
               -\sum_{i=1}^nb_i(K_{\sigma,s}({  x}-{  y})-q({  x}-\eta_i))\\
               &-&
               \sum_{i=1}^nb_iq({  x}-\eta_i).
\end{eqnarray*}
Since $P_l({  y})q({  x}-{  y})\in \mathcal{P}_{l+u}^d(B)$ for fixed
${  x}$, it follows from Lemma \ref{CUBATURE ALG} that with
confidence at least $1-2\exp\{-cn/(u+l)^d\},$ there exists a set of
real numbers $\{w_i\}_{i=1}^n\subset\mathbf R$
 such that
$$
          \int_{B}P_l({y})q({  x}-{ y})d{  y}=\sum_{i=1}^nw_iP_l(\eta_i)q({  x}-\eta_i).
$$
If we set  $a_i=w_iP_l(\eta_i)$,  then
\begin{eqnarray*}
              &&\int_{B}P_l({  y})K_{\sigma,s}({  x}-{  y})d{
              y}-\sum_{i=1}^na_iK_{\sigma,s}({  x}-\eta_i)\\
              &=&
              \int_{B}P_l({  y})(K_{\sigma,s}({ x}-{  y})-q({  x}-{
              y}))d{  y}\\
              &-&
              \sum_{i=1}^nw_iP_l(\eta_i)(K_{\sigma,s}({  x}-\eta_i)-q({  x}-\eta_i))
\end{eqnarray*}
 holds with   confidence at least $1-2\exp\{-cn/(u+l)^d\}$.
Under this circumstance,
\begin{eqnarray*}
                 &&
                 \left\|\int_{B}P_l({  y})K_{\sigma,s}(\cdot-{ y})d{ y}-\sum_{i=1}^na_iK_{\sigma,s}(\cdot-\eta_i)\right\|_{I^d}\\
                  &\leq&
                  \left\|\int_{B}P_l({ y})(K_{\sigma,s}(\cdot-{  y})-q(\cdot-{
                  y}))d{
                  y}\right\|_{I^d}\\
                  &+&
                  \left\|\sum_{i=1}^nw_iP_l(\eta_i)(K_{\sigma,s}(\cdot-\eta_i)-q(\cdot-\eta_i))\right\|_{I^d}
\end{eqnarray*}
To bound the above quantities, denote
$$
           \mathcal L_j(v):=\exp{-\frac{2v}{j^2\sigma^2}}.
$$
Let $\mathcal{T}_u^1([0,(1+a)^2])$ be the set of univariate
algebraic polynomials of degrees not larger than $u$ defined on
$[0,(1+a)^2]$, and set $q_u^j=\arg\min_{q\in
\mathcal{T}_u^1([0,(1+a)^2]^d}\|\mathcal L_j-q\|$, and
$$
           q_u(v):=\sum_{j=1}^s\left(\begin{array}{c} s\\
             j\end{array}\right)\frac1{j^d}\left(\frac{2}{\sigma^2\pi}\right)^{d/2}q_u^j(v).
$$
Then, it follows from (\ref{Jackson1}) that
\begin{eqnarray*}
                   &&
                   \left\|\int_{B}P_l({  y})(K_{\sigma,s}(\cdot-{ y})-q_u((\cdot-{  y})^2))d{
                   y}\right\|_{I^d}\\
                   &\leq&
                   C\|f\|_{I^d}\left\|K_{\sigma,s}({\cdot}-{
                   y})-q_u(({\cdot}-{
                   y}\right)^2)\|_{I^d}\\
                   &\leq&
                   C \|f\|_{I^d}\sum_{j=1}^r\left(\begin{array}{c} r\\
             j\end{array}\right)\frac1{j^d}\left(\frac{2}{\sigma^2\pi}\right)^{d/2}\\
             &\times&
             \inf_{q\in \mathcal{T}_u^1([0,(1+a)^2])}\|\mathcal L_j-q\|.
\end{eqnarray*}
On the other hand, since $\sum_{i=1}^n|w_i|\leq
\sqrt{n\sum_{i=1}^n|w_i|^2}\leq C$, we  also obtain
\begin{eqnarray*}
                   &&
                   \left\|\sum_{i=1}^nw_iP_l(\eta_i)(K_{\sigma,s}(\cdot-\eta_i)-q_u((\cdot-\eta_i)^2))\right\|\\
                   &\leq&
                    C \|f\|_{I_d}\sum_{j=1}^s\left(\begin{array}{c} s\\
             j\end{array}\right)\frac1{j^d}\left(\frac{2}{\sigma^2\pi}\right)^{d/2}\\
             &\times&
             \inf_{q\in \mathcal{T}_u^1([0,(1+a)^2])}\|\mathcal L_j-q\|.
\end{eqnarray*}
Thus, the only thing remainder  is to bound $
        \int_{B}(F({ y})-P_l({  y}))K_{\sigma,s}({  x}-{
             y})d{  y}.
$ It follows from (\ref{Jackson}) that
\begin{eqnarray*}
            &&
            \left\|\int_{B}(F({  y})-P_l({  y}))K_{\sigma,s}({  x}-{
             y})d{  y}\right\|\\
             &\leq&
              E_l(F)\times\int_{B}K_{\sigma,s}({  x}-{
             y})d{\bf y}
             \leq C_s \omega_{s,\mathbf R^d}(F,1/l),
\end{eqnarray*}
where we use the fact \cite{Eberts2011}
$$
               \int_{B}K_{\sigma,s}({  x}-{
             y})d{\bf y}\leq 1
$$
and the known Jackson inequality \cite{DeVore1993} in the last
inequality.
 All above together with Lemma \ref{TOPIC} yields that
\begin{eqnarray*}
      &&\inf_{g_n\in\mathcal G_n}\|K_{\sigma,s}\ast F-g_n\|\leq C_s
      \omega_{s,I^d}(f,1/l)\\
      &+&C_s
      a\|f\|\sigma^d
      +
      C \|f\|\sum_{j=1}^s\left(\begin{array}{c} s\\
             j\end{array}\right)\frac1{j^d}\left(\frac{2}{\sigma^2\pi}\right)^{d/2}\\
             &\times&
             \inf_{q\in \mathcal{T}_u^1([0,(1+a)^2])}\|\mathcal L_j-q\|
\end{eqnarray*}
holds with confidence at least $1-2\exp\{-cn/(u+l)^d\}.$
Furthermore, it is straightforward to check, using the power series
\cite[P.136]{Mhaskar1999} for $\exp\{-\frac{2v}{j^2\sigma^2}\}$ that
\begin{eqnarray*}
                &&\sum_{j=1}^s\left(\begin{array}{c} s\\
             j\end{array}\right)\frac1{j^d}\left(\frac{2}{\sigma^2\pi}\right)^{d/2}\inf_{q\in \mathcal{T}_u^1([0,(1+a)^2])}\|\mathcal
             L_j-q\|\\
             &\leq& C_s \sigma^{-d} \frac{2^u}{ u!\sigma^2}.
\end{eqnarray*}
Thus, the proof of Lemma 9 is completed.
\end{IEEEproof}

By the help of the above nine lemmas, we can proceed the proof of
Theorem \ref{APPROXIMATION ERROR} as follows.

\begin{IEEEproof}[Proof of Theorem \ref{APPROXIMATION ERROR}]
 Since
$$
             \inf_{g_n\in\mathcal H_{\sigma,s,n}}\| f-g_n  \|_{I^d}\leq \|f-  K_{\sigma,s}\ast F\|_{I^d}+\|K_{\sigma,s}\ast F-g_n\|_{I^d},
$$
Setting $\sigma=l^{-1/2}$, it follows from Lemma 2 and Lemma 9 that
\begin{eqnarray*}
          &&\inf_{g_n\in\mathcal H_{\sigma,s,n}}\| f-g_n  \|_{I^d}\leq
          C_s\left(\omega_{s,I^d}(f,l^{-1/2})+
      a\|f\|\sigma^d\right.\\
      &+&
     \left.\sigma^{-d} \frac{(s^2\sigma^2)^u}{2^u u!}\right)
\end{eqnarray*}
holds with confidence at least
 $1-2\exp\{-cn/(u+l)^d\}.$
 By the Stirling's formula, it is easy to check that
$$
             \sigma^{-d} \frac{(s^2\sigma^2)^u}{2^u u!}\leq
             Cu^{d}\frac{(u/2)^u}{2^uu!}\leq C\frac{u^d}{(2d)^u}\leq
             Cl^{-d/2}
$$
with $u=2dl$. Therefore, we obtain
$$
      \inf_{g_n\in\mathcal H_{\sigma,s,n}}\|f-g_n\|\leq C_s
      \left(\omega_{s,I^d}(f,l^{-1/2})+
      a\|f\|l^{-d/2}\right),
$$
with confidence at least $1-2\exp\{-cn/l^d\}.$ Therefore, Theorem
\ref{APPROXIMATION ERROR} follows by noticing  $\sigma=1/\sqrt{l}$.
\end{IEEEproof}

\subsection{Proof of Proposition \ref{LOWER BOUND APPROXIMATION}}
To prove Proposition \ref{LOWER BOUND APPROXIMATION}, we need the
following two lemmas, the first one concerning  Bernstein inequality
for $\mathcal H_{\sigma,s,n}$  can be easily deduced from \cite[eqs
(3.1)]{Erdelyi2006}.

\begin{lemma}\label{BERNSTEIN FOR GAUSSIAN}
Let $d=1$, $s=1$,   and $\sigma\geq n^{-1/2}$. Then, for arbitrary
$g_n\in\mathcal H_{\sigma,s,n}$, there holds
$$
          \|g_n'\|_{[0,1]}\leq Cn^{1/2}\|g_n\|_{[0,1]},
$$
where $C$ is an absolute constant.
\end{lemma}

By the help of the Bernstein inequality, the standard method in
approximation theory \cite[Chap. 7]{DeVore1993}  yields the
following Lemma \ref{INVERSE TH}.

\begin{lemma}\label{INVERSE TH}
Let $d=1$, $s=1$, $r\in\mathbf N$,  $\sigma\geq n^{-1/2}$ and $f\in
C(I^1)$. If
$$
          \sum_{n=1}^\infty
          n^{r/2-1}\mbox{dist}(f,\mathcal
          H_{\sigma,1,n})<\infty,
$$
then $f\in\mathcal F^{r}$, where $\mbox{dist}(f,\mathcal
               H_{\sigma,1,n})=\inf_{g\in\mathcal
               H_{\sigma,1,n}}\|f-g\|_{I^1}$.
\end{lemma}

\begin{IEEEproof}
Let $g_n:=\arg\inf_{g\in\mathcal
               H_{\sigma,1,n}}\|f-g\|_{I^1}$.
For  arbitrary $n\in\mathbf N$, set $n_0$ such that
$$
           2^{n_0}\leq n\leq 2^{n_0+1}.
$$
 It is easy to see that
$$
          \sum_{n=1}^\infty
          n^{r/2-1}\mbox{dist}(f,\mathcal H_{\sigma,1,n})<\infty,
$$
 implies $\mbox{dist}(f,\mathcal H_{\sigma,1,n})\rightarrow 0$ in $C(I^1)$. Indeed,
 if it does not hold, then there exists an absolute constant $C$
such that $\mbox{dist}(f,\mathcal H_{\sigma,1,n})\geq C>0$.
Therefore,
$$
           C\sum_{n=1}^{\infty}n^{-1}<\sum_{n=1}^{\infty}n^{\frac{r}2-1}\mbox{dist}(f,\mathcal
           H_{\sigma,1,n})
          <\infty,
$$
which is impossible.  So we have
\begin{equation}\label{expansion}
                 f-g_{2^{n_0}}=\sum_{j=n_0}^\infty g_{2^{j+1}}-g_{2^j}.
\end{equation}
By Lemma \ref{BERNSTEIN FOR GAUSSIAN}, we then have
$$
           \|g_{2^{j+1}}'-g_{2^j}'\|_{I^1} \leq
              C2^{(j+1)r/2}\mbox{dist}(f,\mathcal H_{\sigma,1,2^{j}}).
$$
Then direct computation yields that
\begin{eqnarray*}
                  \|g_{2^{j+1}}'-g_{2^j}'\|_{I^1}
                  &\leq&
                C \sum_{j=1}^{\infty}\sum_{k=2^{j-1}+1}^{2^j}k^{r/2-1}\mbox{dist}(f,\mathcal H_{\sigma,1,k
                })\\
              &\leq&
               C \sum_{k=1}^{\infty}k^{r/2-1}\mbox{dist}(f,\mathcal H_{\sigma,1,k })
               <\infty.
\end{eqnarray*}
So $\{g_{2^j}\}$ is the Cauchy sequence of $\mathcal F^{r}$.
Differentiating (\ref{expansion}), we have
$$
           f'-g'_{2^{n_0}}= \sum_{j=n_0}^{\infty}g_{2^{j+1}}-g_{2^j},
$$
Since $\{g_{2^j}\}$ is the Cauchy sequence of $\mathcal F^{r}$,  we
have $f'-g'_{2^{n_0}}\rightarrow 0$ when $n_0\rightarrow\infty$,
which implies $f\in \mathcal F^{r}$.
\end{IEEEproof}

Now we continue the proof of Proposition \ref{LOWER BOUND
APPROXIMATION}.

\begin{IEEEproof}[Proof of Proposition \ref{LOWER BOUND
APPROXIMATION}] Let $\varepsilon\in(0,1),$ and $r=1-\varepsilon$. It
is obvious that there exists a function $h_r$ satisfying $h_r\in
\mathcal F^r$ and $h_r\notin F^{r'}$ with $r'>r$. Assume
$$
           \inf_{g\in\mathcal H_{\sigma,1,n}}\|f-g\|\leq
           Cn^{-r/2-\varepsilon}
$$
holds for all $f\in\mathcal F^r$, where $C$ is a constant
independent of $n$. Then,
$$
          \inf_{g\in\mathcal H_{\sigma,1,n}}\|h_r-g\|\leq
           Cn^{-r/2-\varepsilon}.
$$
Then,
$$
           \sum_{n=1}^\infty n^{1/2-1}\mbox{dist}(h_r,\mathcal
           H_{\sigma,1,n})=\sum_{n=1}^\infty
           n^{-1-\varepsilon/2}<\infty.
$$
Therefore, it follows from Lemma \ref{INVERSE TH} that $h_r\in
\mathcal F^1$, which is impossible. Hence,
$$
          \sup_{f\in\mathcal F^r}\inf_{g\in\mathcal
            H_{\sigma,1,n}}\|f-g\|\geq
           Cn^{-r/2-\varepsilon}.
$$
 This together with Theorem \ref{APPROXIMATION ERROR} finishes the proof of Proposition \ref{LOWER BOUND APPROXIMATION}.
\end{IEEEproof}

\subsection{Proof of Theorem \ref{THEOREM LEARNG}}

The main tool to prove Theorem \ref{THEOREM LEARNG} is the following
Lemma \ref{CONCENTRATION}, which can be found in
\cite[Chap.11]{Gyorfi2002}.

\begin{lemma}\label{CONCENTRATION}
 Let  $f_{{\bf z},\sigma,s,n}$ be defined as in (\ref{Gaussian ELM Estimator}).
Then
\begin{eqnarray}\label{oracle}
                 &&E_{\rho^m}\|\pi_M f_{{\bf z},\sigma,s,n}
                 -f_\rho\|_\rho^2
                 \leq
                 CM^2\frac{(\log m+1)n}{m}\nonumber\\
                 &+&
                 8 \inf_{f \in \mathcal{H}_{\sigma,s,n}}\int_X |f(x)- f_\rho(x)|^2
                 d\rho_X
\end{eqnarray}
for some universal constant C.
\end{lemma}

Now, we use Proposition \ref{APPROXIMATION ERROR} and Lemma
\ref{CONCENTRATION} to prove Theorem \ref{THEOREM LEARNG}.

\begin{IEEEproof}[Proof of Theorem \ref{THEOREM LEARNG}]
Since $\mathcal H_{\sigma,s,n}$ is a $n$-dimensional linear space,
then Lemma \ref{CONCENTRATION} yields that
\begin{eqnarray*}
                 &&E_{\rho^m}\|\pi_M
          f_{{\bf z},\sigma,s,n} -f_\rho\|_\rho^2
                 \leq
                 CM^2\frac{(\log m+1)n}{m}\\
                 &+&
                 8 \inf_{f \in \mathcal{H}_{\sigma,s,n}}\int_X |f(x)- f_\rho(x)|^2
                 d\rho.
\end{eqnarray*}
Therefore, it suffices to bound
$$
                    \inf_{f \in \mathcal{H}_{\sigma,s,n}}\int_X |f(x)-
                    f_\rho(x)|^2\leq \inf_{f \in
                    \mathcal{H}_{\sigma,s,n}}\|f-f_\rho\|_X^2.
$$
From Theorem \ref{APPROXIMATION ERROR}, it follows that
 $$
                      \inf_{g \in
                    \mathcal{H}_{\sigma,s,n}}\|g-f_\rho\|_X\leq C\left(\omega_{s,  I^d}(f_\rho,\sigma)+
                       \|f_\rho\|\sigma^d\right)
 $$
holds with probability at least $1-2\exp\{-cn\sigma^{2d}\}$. Noting
$r\leq s$ and $f_\rho\in\mathcal F^r$, with probability at least
$1-2\exp\{-cn\sigma^{2d}\}$, there holds
$$
                      \inf_{f \in
                    \mathcal{H}_{\sigma,s,n}}\|f-f_\rho\|^2_X\leq C\left(\sigma^{2r}+
                     \sigma^{2d}\right).
 $$
Setting $\sigma=n^{(-1+\varepsilon)/(2d)}$, we observe that with
probability at least $1-2\exp\{-n^{\varepsilon}\}$, there holds
$$
                          \inf_{f \in
                    \mathcal{H}_{\sigma,s,n}}\|f-f_\rho\|^2_X\leq
                    C\left(n^{-r/d+r\varepsilon/d}+n^{-1+\varepsilon}\right).
$$
Finally, choosing $n=\left[m^{\frac{d}{r+d}}\right]$, we obtain that
with probability at least $1-2\exp\{-n^{\varepsilon}\}$, there holds
$$
           E_{\rho^m}\|\pi_M
          f_{{\bf z},\sigma,s,n} -f_\rho\|_\rho^2
          \leq C\left(m^{-\frac{(1-\varepsilon)r}{r+d}}\log m
          +m^{-\frac{d(1-\varepsilon)}{r+d}}\right).
$$
This finishes the proof of Theorem \ref{THEOREM LEARNG}.
\end{IEEEproof}

\subsection{Proof of Proposition \ref{LOWER BOUND LEARNING}}
To prove Proposition \ref{LOWER BOUND LEARNING}, we need the
following three lemmas. The first one is the interpolation theorem
of linear functionals, which can be found in
\cite[P.385]{Borwein1995}.

\begin{lemma}\label{FUNCTIONAL INTERPOLATION}
  Let $C(Q)$ be the set of real valued
continuous functions on the compact Hausdorff space $Q$. Let $S$ be
an $n$-dimensional linear subspace of $C(Q)$ over $\mathbf{R}.$ Let
$L\neq 0$ be a real-valued linear functional on $S$. Then there
exist points $x_1,x_2,\dots, x_r$ in $Q$ and nonzero real numbers
$a_1, a_2,\dots, a_r$, where $1\leq r\leq n$, such that
$$
           L(s)=\sum_{i=1}^ra_is(x_i),\ \ \ s\in S
$$
 and
$$
             \|L\|=\sup \bigl\{|L(s)|:s\in S,\ \|s\|_Q \leq
               1\bigr\}=\sum_{i=1}^r|a_i|.
$$
\end{lemma}

By using Lemmas \ref{FUNCTIONAL INTERPOLATION} and \ref{BERNSTEIN
FOR GAUSSIAN}, we can obtain the following  Bernstein inequality for
ELM with Gaussian kernel in the metric of $L^2_{\rho_X}$.

\begin{lemma}\label{BERNSTEIN FOR GAUSSIAN11111}
Let $d=1$, $s=1$,   and $\sigma\geq n^{-1/2}$. Then, for arbitrary
$g_n\in\mathcal H_{\sigma,s,n}$, there holds
$$
          \|g_n'\|_\rho\leq Cn^{1/2}\|g_n\|_\rho,
$$
where $C$ is an absolute constant.
\end{lemma}

\begin{IEEEproof}
  We apply Lemma \ref{FUNCTIONAL INTERPOLATION}
with $Q=[1/2,1],\ S=\mathcal H_{\sigma,s,n},$ and
$L(s)=s^{\prime}(1).$ It follows from Lemma \ref{BERNSTEIN FOR
GAUSSIAN} that
\begin{equation}\label{tool to prove proposition3}
            \|L\|=|s^{\prime}(1)|\leq C n^{1/2}|s(1)|= C_1n^{1/2}.
\end{equation}
We deduce that there are $v_1,v_2,\dots, v_r$ in $[1/2,1]$ and
$a_1,a_2,\dots, a_r\in I^1$ so that for every $s\in \mathcal
H_{\sigma,s,n}$,
$$
             \frac{|s'(1)|}{C_1n^{1/2}}=\frac{|\sum_{i=1}^ra_is(v_i)|}{C_1n^{1/2}}
             \leq \sum^r_{i=1}\left|\frac{a_i}{C_1n^{1/2}}\right||s(v_i)|
$$
with $1\leq r\leq n$. By (\ref{tool to prove proposition3}) we have
$$
             \sum_{i=1}^r\left|\frac{a_i}{C_1n^{1/2}}\right|\leq 1.
$$
So there is a sequence of numbers $\{c_i\}$ with $\sum_{i=1}^r
|c_i|=1$ such that
$$
              \frac{|s'(1)|}{C_1n^{1/2}}\leq\sum_{i=1}^r|c_i||s(v_i)|.
$$
Now let $\phi:[0,\infty)\rightarrow [0,\infty)$ be a nondecreasing
convex function. Using monotonicity and convexity, we have
$$
           \phi\left(\frac{|s'(1)|}{C_1n^{1/2}}\right )
             \leq
           \phi(\sum_{i=1}^r|c_is(v_i)|)
            \leq \sum_{i=1}^r|c_i|\phi(|s(v_i)|).
 $$
Applying this inequality with $s(t)= g_n(t+u-1)\in \mathcal
H_{\sigma,s,n}$, we get
$$
            \phi\left(\frac{|g_n^{\prime}(u)|}{C_1n^{1/2}}\right ) \leq
             \sum_{i=1}^r|c_i|\phi(|P(v_i+u-b)|)
$$
for every $P\in \mathcal H_{\sigma,s,n}$ and $u\in [1/2,1]$. Since
$x_i\in[1/2,1]$ and $u\in [1/2,1]$, then $v_i+u-1\in [0,1]$ for each
$i=1,2,\dots, r.$ Integrating on the interval $[1/2,1]$ with respect
to $u$, we obtain
\begin{eqnarray*}
                &&\int_{1/2}^1\phi\left(\frac{|g_n'(u)|}{C_1n^{1/2}}\right)d\rho_X(u)\\
                 &\leq&
                  \sum_{i=1}^r\int_{1/2}^1|c_i|\phi(|g_n(v_i+u-1)|)d\rho_X(u) \\
                &\leq& \sum_{i=1}^r\int_0^1|c_i|\phi(|g_n(t)|)d\rho_X(t) \leq
              \int_0^1\phi(|g_n(t)|)dt,
\end{eqnarray*}
in which $\sum_{i=1}^r|c_i|=1$ has been used.

It can be shown exactly in the same way that
$$
          \int_0^{1/2}\phi\left(\frac{|g_n'(u)|}{C_1\lambda_n}\right)d\rho_X(u) \leq
            \int_0^1\phi(|g_n(t)|)d\rho_X(t).
$$
Combining the last two inequalities and choosing $\phi(x)=x^2,$ we
finish the proof of Lemma \ref{BERNSTEIN FOR GAUSSIAN11111}.
\end{IEEEproof}

Using almost the same method as that in the proof of Lemma
\ref{INVERSE TH}, the following Lemma \ref{INVERSE TH111} can be
deduced directly from Lemma \ref{BERNSTEIN FOR GAUSSIAN11111}

\begin{lemma}\label{INVERSE TH111}
Let $d=1$, $s=1$, $r\in\mathbf N$,  $\sigma\geq n^{-1/2}$ and $f\in
C(I^1)$. If
$$
          \sum_{n=1}^\infty
          n^{r/2-1}\mbox{dist}(f,\mathcal
          H_{\sigma,1,n})_\rho<\infty,
$$
then $f\in\mathcal F^{r}$, where $\mbox{dist}(f,\mathcal
               H_{\sigma,1,n})_\rho=\inf_{g\in\mathcal
               H_{\sigma,1,n}}\|f-g\|_{\rho}$.
\end{lemma}

Now, we proceed the proof of Proposition \ref{LOWER BOUND LEARNING}.
\begin{IEEEproof}[Proof of Proposition \ref{LOWER BOUND LEARNING}]
With the help of the above lemmas, we can use the almost same method
as that in the proof of Proposition \ref{LOWER BOUND APPROXIMATION}
to obtain
$$
          \sup_{f\in\mathcal F^r}\inf_{g\in\mathcal
            H_{\sigma,1,n}}\|f-g\|_\rho\geq
           Cn^{-r/2-\varepsilon}.
$$
Then,
 Proposition \ref{LOWER BOUND LEARNING}
 can be deduced from
 the above inequality by using the
 conditions,
$\sigma=m^{ \frac{-1+\varepsilon}{2+2r}}$ and
$n=\left[m^{\frac{1}{1+r}}\right]$.
\end{IEEEproof}

\subsection{Proof of Theorem \ref{REGULARIZATION THEOREM}}
To prove Theorem \ref{REGULARIZATION THEOREM}, we need the following
concepts and lemmas.  Let $(\mathcal M,\tilde{d})$ be a
pseudo-metric space and $T\subset\mathcal M$ a subset. For every
$\varepsilon>0$, the covering number $\mathcal
N(T,\varepsilon,\tilde{d})$ of $T$ with respect to $\varepsilon$ and
$\tilde{d}$ is defined as the minimal number of balls of radius
$\varepsilon$ whose union covers $T$, that is,
$$
                 \mathcal N(T,\varepsilon,\tilde{d}):=\min\left\{l\in\mathbf
                 N: T\subset\bigcup_{j=1}^lB(t_j,\varepsilon)\right\}
$$
for some $\{t_j\}_{j=1}^l\subset\mathcal M$, where
$B(t_j,\varepsilon)=\{t\in\mathcal
M:\tilde{d}(t,t_j)\leq\varepsilon\}$. The $l^2$-empirical covering
number \cite{Sun2011} of a function set is defined by means of the
normalized $l^2$-metric $\tilde{d}_2$ on the Euclidean space
$\mathbf R^d$ given in with $
                  \tilde{d}_2({\bf
                  a,b})=\left(\frac1m\sum_{i=1}^m|a_i-b_i|^2\right)^\frac12
$
 for  ${\bf a}=(a_i)_{i=1}^m, {\bf
                  b}=(b_i)_{i=1}^m\in\mathbf R^m.$
\begin{definition}
 Let $\mathcal G$ be a set of functions on $X$,
${\bf x}=(x_i)_{i=1}^m$,  and
$$
            \mathcal G|_{\bf x}:=\{(f(x_i))_{i=1}^m:f\in\mathcal G\}\subset R^m.
$$
 Set $\mathcal
N_{2,{\bf x}}(\mathcal G,\varepsilon)=\mathcal N(\mathcal G|_{\bf
x},\varepsilon,\tilde{d}_2)$. The $l^2$-empirical covering number of
$\mathcal G$ is defined by
$$
                 \mathcal N_2(\mathcal
                 F,\varepsilon):=\sup_{m\in\mathbf N}\sup_{{\bf
                 x}\in S^m}\mathcal N_{2,{\bf x}}(\mathcal
                 G,\varepsilon),\ \ \varepsilon>0.
$$
\end{definition}

Let $H_\sigma$ be the reproducing kernel Hilbert space of
$K_{\sigma,s}$ \cite{Steinwart2007} and $B_{H_\sigma}$ be the unit
ball in $H_\sigma$. The following Lemmas \ref{COVERINGNUMBER} and
\ref{CONCENTRATION L2}
  can be easily deduced from \cite[Theorem 2.1]{Steinwart2007}
and \cite{Sun2011}, respectively.

\begin{lemma}\label{COVERINGNUMBER}
Let $0<\sigma\leq 1$, $X\subset \mathbf R^d$ be a compact subset
with nonempty interior. Then for all $0<p\leq 2$ and all $\nu>0$,
there exists a constant $C_{p,\nu,d,s}>0$ independent of $\sigma$
such that for all $\varepsilon>0$, we have
$$
          \log \mathcal N_2(B_{H_\sigma},\varepsilon)\leq
          C_{p,\mu,d,s}\sigma^{(p/2-1)(1+\nu)d}\varepsilon^{-p}.
$$
\end{lemma}

\begin{lemma}\label{CONCENTRATION L2}
  Let $\mathcal F$ be a
class of measurable functions on $Z$. Assume that there are
constants $B,c>0$ and $\alpha\in[0,1]$ such that $\|f\|_\infty\leq
B$ and $\mathbf Ef^2\leq c(\mathbf E f)^\alpha$ for every
$f\in\mathcal F.$ If for some $a>0$ and $p\in(0,2)$,
\begin{equation}\label{condition}
                  \log\mathcal N_2(\mathcal F,\varepsilon)\leq
                  a\varepsilon^{-p},\ \ \forall\varepsilon>0,
\end{equation}
then there exists a constant $c_p'$ depending only on $p$ such that
for any $t>0$, with probability at least $1-e^{-t}$, there holds
\begin{eqnarray}\label{lemma3}
                 &&\mathbf Ef-\frac1m\sum_{i=1}^mf(z_i)
                 \leq
                 \frac12\eta^{1-\alpha}(\mathbf
                 Ef)^\alpha+c_p'\eta\nonumber\\
                 &+&
                 2\left(\frac{ct}{m}\right)^\frac1{2-\alpha}
                 +
                 \frac{18Bt}{m},\
                 \forall f\in\mathcal F,
\end{eqnarray}
where
$$
               \eta:=\max\left\{c^\frac{2-p}{4-2\alpha+p\alpha}\left(\frac{a}m\right)^\frac2{4-2\alpha+p\alpha},
               B^\frac{2-p}{2+p}\left(\frac{a}m\right)^\frac2{2+p}\right\}.
$$
\end{lemma}

The next lemma states a variant of Lemma \ref{BERNSTEIN}, which can
be found in \cite{Shi2011}

\begin{lemma}\label{BERNSTEIN1}
 Let $\xi$ be a random variable on a probability space
$Z$ with variance $\gamma^2$ satisfying $|\xi-\mathbf E\xi|\leq
M_\xi$ for some constant $M_\xi$. Then for any $0<\delta<1$, with
confidence $1-\delta$, we have
$$
             \frac1m\sum_{i=1}^m\xi(z_i)-\mathbf
             E\xi\leq\frac{2M_\xi\log\frac1\delta}{3m}+\sqrt{\frac{2\sigma^2\log\frac1\delta}{m}}.
$$
\end{lemma}

From the proof of Lemma \ref{ESTIMATOR1}, we can also deduce the
following Lemma \ref{R APPROX ERROR}

\begin{lemma}\label{R APPROX ERROR}
Let $d,s,n\in\mathbf N$. Then with confidence at least
$1-2\exp\{-cn\sigma^{2d}\}$, there exists a $f_0\in\mathcal
H_{\sigma,s,n}$ such that
$$
             \|f_\rho-f_0\|_\rho^2+\lambda\Omega(f_0) \leq
            C(\omega_{s,I^d}(f_\rho,\sigma)^2+\sigma^{2d}+\lambda/n),
$$
where $C$ is a constant depending only on $d$, $s$ and $M$.
\end{lemma}

\begin{IEEEproof}
Let
$$
            f_0=\sum_{i=1}^na_iK_{\sigma,s}(x-\eta_i)=\sum_{i=1}^nw_iP_l(\eta_i)K_{\sigma,s}(x-\eta_i),
$$
where $\{w_i\}_{i=1}^n$ and $P_l$ are the same as those in the proof
of Lemma \ref{ESTIMATOR1}. Then, it has already been proved that
$$
              \|f_\rho-f_0\|_\rho\leq
              C(\omega_{s,I^d}(f_\rho,\sigma)+\sigma^{d}).
$$
Furthermore, it can be deduced from Lemma \ref{CUBATURE ALG} and
(\ref{Jackson1}) by taking $f=f_\rho$ that
$$
        \Omega(f_0)=\sum_{i=1}^n|w_i|^2|P_l(\eta_i)|^2\leq\|f_\rho\|^2\sum_{i=1}^n|w_i|^2\leq
        C/n.
$$
This finishes the proof of Lemma \ref{R APPROX ERROR}.
\end{IEEEproof}

Now we proceed the proof of Theorem \ref{REGULARIZATION THEOREM}.

\begin{IEEEproof}[Proof of Theorem \ref{REGULARIZATION THEOREM}]
Let $f_{{\bf z}, \sigma,s,\lambda,n}$  and $f_0$ be defined as in
(\ref{regularization}) and Lemma \ref{R APPROX ERROR}, respectively.
Define
$$
         \mathcal D:=\mathcal E(f_0)-\mathcal
            E(f_\rho)+\lambda\Omega(f_0)
$$
and
$$
        \mathcal S:=\mathcal
        E_{\bf z}(f_0)-\mathcal E(f_0)+\mathcal E(\pi_Mf_{{\bf z},\sigma,s,\lambda,n})-\mathcal
        E_{\bf z}(\pi_Mf_{{\bf z},\sigma,s,\lambda,n}),
$$
where $\mathcal E_{\bf
               z}(f)=\frac1m\sum_{i=1}^m(y_i-f(x_i))^2$.
Then, it is easy to check that
\begin{equation}\label{error decomposition}
         \mathcal E(\pi_Mf_{{\bf z},\sigma,s,\lambda,n})-\mathcal E(f_\rho)
         \leq
        \mathcal D+\mathcal S.
\end{equation}
As $f_\rho\in \mathcal F^r$, it follows from Lemma \ref{R APPROX
ERROR} that with confidence at least $1-2\exp\{-cn\sigma^{2d}\}$
(with respect to $\mu^n$), there holds
\begin{equation}\label{r approximation error estimate}
                  \mathcal D\leq
                  C(\sigma^{2r}+\sigma^{2d}+\lambda/n).
\end{equation}
Upon using the short hand notations
$$
               \mathcal S_1:=\{\mathcal E_{\bf
               z}(f_0)-\mathcal E_{\bf
               z}(f_\rho)\}-\{\mathcal E(f_0)-\mathcal
               E(f_\rho)\}
$$
and
$$
               \mathcal S_2:=\{\mathcal E(\pi_Mf_{{\bf z},\sigma,s,\lambda,n})-\mathcal E(f_\rho)\}-\{\mathcal E_{\bf
               z}(\pi_Mf_{{\bf z},\sigma,s,\lambda,n})-\mathcal E_{\bf z}(f_\rho)\},
$$
we have
\begin{equation}\label{sample decomposition}
            \mathcal S=\mathcal S_1+\mathcal
            S_2.
\end{equation}

We first turn to bound $\mathcal S_1$.
 Let the random variable $\xi$ on $Z$ be defined by
$$
               \xi({\bf z})=(y-f_0(x))^2-(y-f_\rho(x))^2 \quad {\bf z}=(x,y)\in Z.
$$
Since $|f_\rho(x)|\leq M$ and
$$
       |f_0|\leq\sum_{i=1}^n|w_i||P_l(\eta_i)||K_{\sigma,s}(\eta_i,x)|\leq
       \|f_\rho\|\sum_{i=1}^n|w_i|\leq CM
$$
hold almost everywhere, we have
\begin{eqnarray*}
          |\xi({\bf z})|
          &=&
          (f_\rho(x)-f_0(x))(2y-f_0(x)-f_\rho(x))\\
          &\leq&
          (M+CM)(3M+CM)
          \leq
           M_\xi:=(3M+CM)^2
\end{eqnarray*}
and almost surely
$$
            |\xi-\mathbf E\xi|\leq 2M_\xi.
$$
Moreover, we have
\begin{eqnarray*}
            E(\xi^2)
            &=&
            \int_Z(f_0(x)+f_\rho(x)-2y)^2(f_0(x)-f_\rho(x))^2d\rho\\
            &\leq&
             M_\xi\|f_\rho-f_0\|^2_\rho,
\end{eqnarray*}
which implies that the variance $\gamma^2$ of $\xi$ can be bounded
as $\gamma^2\leq E(\xi^2)\leq M_\xi\mathcal D.$ Now applying Lemma
\ref{BERNSTEIN1}, we obtain
\begin{eqnarray}\label{bound s1}
       \mathcal
       S_1
       &\leq&
       \frac{4M_\xi\log\frac2\delta}{3m}+\sqrt{\frac{2M_\xi\mathcal
       D \log\frac{2}{\delta}}{m}}\nonumber\\
       &\leq&
       \frac{7(3M+CM)^2\log\frac2\delta}{3m}+\frac12\mathcal D
\end{eqnarray}
holds with confidence $1-\frac\delta2$ (with respect to $\rho^m$).

To bound $\mathcal S_2$, we need apply Lemma \ref{CONCENTRATION L2}
to the set $\mathcal G_R$, where
$$
        \mathcal G_R:=\left\{(y-\pi_Mf(x))^2-(y-f_\rho(x))^2:f\in
       \mathcal B_R\right\}
$$
and
$$
        \mathcal B_R:=\left\{f=\sum_{i=1}^nb_iK_{\sigma,s}(\eta_i,x):
        \sum_{i=1}^n|b_i|^2\leq R\right\}.
$$
 Each function
$g\in\mathcal G_R$ has the form
$$
         g(z)=(y-\pi_Mf(x))^2-(y-f_\rho(x))^2, \quad f\in \mathcal
         B_R
$$
and is automatically a function on $Z$. Hence
$$
           \mathbf Eg=\mathcal E(f)-\mathcal
           E(f_\rho)=\|\pi_Mf-f_\rho\|_\rho^2
$$
and
$$
           \frac1m\sum_{i=1}^mg(z_i)=\mathcal E_{\bf z}(\pi_Mf)-\mathcal
           E_{\bf z}(f_\rho),
$$
where $z_i:=(x_i,y_i)$. Observe that
$$
           g(z)=(\pi_Mf(x)-f_\rho(x))((\pi_Mf(x)-y)+(f_\rho(x)-y)).
$$
Therefore,
$$
            |g(z)|\leq 8M^2
$$
and
\begin{eqnarray*}
           \mathbf
           Eg^2
           &=&
           \int_Z(2y-\pi_Mf(x)-f_\rho(x))^2(\pi_Mf(x)-f_\rho(x))^2d\rho\\
           &\leq&
           16M^2\mathbf Eg.
\end{eqnarray*}
For $g_1,g_2\in\mathcal F_{R_q}$ and arbitrary $m\in\mathbf N$, we
have
\begin{eqnarray*}
        &&\left(\frac1m\sum_{i=1}^m(g_1(z_i)-g_2(z_i))^2\right)^{1/2}\\
        &\leq&
        \left(\frac{4M}m\sum_{i=1}^m(f_1(x_i)-f_2(x_i))^2\right)^{1/2}.
\end{eqnarray*}
It follows that
$$
         \mathcal N_{2,{\bf z}}(\mathcal G_{R},\varepsilon)
         \leq
          \mathcal N_{2,{\bf
         x}}\left(
         \mathcal B_R,\frac\varepsilon{4M
         }\right)
         \leq
         \mathcal N_{2,{\bf x}}\left(
         \mathcal B_1,\frac\varepsilon{4MR}\right),
$$
which together with  Lemma \ref{COVERINGNUMBER} implies
$$
            \log \mathcal N_{2,{\bf z}}(\mathcal G_R,\varepsilon)\leq
           C_{p,\mu,d}\sigma^{\frac{p-2}2(1+\nu)d}(4MR)^p\varepsilon^{-p}.
$$
By Lemma \ref{CONCENTRATION L2} with $B=c=16M^2$, $\alpha=1$ and
$a=C_{p,\mu,d}\sigma^{\frac{p-2}2(1+\nu)d}(4MR)^p$, we know that for
any $\delta\in (0,1),$ with confidence $1-\frac\delta2,$ there
exists a constant $C$ depending only on $d$  such that for all
$g\in\mathcal G_R$
$$
        \mathbf Eg-\frac1m\sum_{i=1}^mg(z_i)
        \leq
        \frac12\mathbf
        Eg+C\eta+C(M+1)^2\frac{\log(4/\delta)}{m}.
$$
Here
$$
             \eta=\{16M^2\}^{\frac{2-p}{2+p}}C_{p,\nu,d}^{\frac2{2+p}}m^{-\frac{2}{2+p}}\sigma^{\frac{p-2}2(1+\nu)d\frac{2}{2+p}}R^{\frac{2p}{2+p}}.
$$
Hence, we obtain
\begin{eqnarray*}
           &&\mathbf
           Eg-\frac1m\sum_{i=1}^mg(z_i)
           \leq
           \frac12\mathbf
           Eg
           +
           \{16(M+1)^2\}^{\frac{2-p}{2+p}}C_{p,\nu,d}^{\frac2{2+p}}
           \\
           &\times&
           m^{-\frac{2}{2+p}}
           \sigma^{\frac{p-2}2(1+\nu)d\frac{2}{2+p}}R^{\frac{2p}{2+p}}\log\frac{4}{\delta}.
\end{eqnarray*}
Now we turn to estimate $R$.
 It follows form the definition of $f_{{\bf z},\sigma,s,\lambda,n}$ that
$$
             \lambda\Omega(f_{{\bf z},\sigma,s,\lambda,n})\leq\mathcal E_{\bf
             z}(0)+\lambda\cdot0\leq M^2.
$$
Thus, we obtain that for arbitrary $0<p\leq 2$ and arbitrary
$\nu>0$, there exists a
 constant $C$ depending only on $d$, $\nu$, $p$ and $M$ such that
\begin{eqnarray}\label{equation}\label{bound s2}
        \mathcal S_2
        &\leq&
        \frac12\{\mathcal E(\pi_Mf_{{\bf z},\sigma,s\lambda,n})-\mathcal
        E(f_\rho)\}\nonumber\\
        &+&
        C \log\frac4\delta
        m^{-\frac{2}{2+p}}\sigma^{\frac{(p-2)(1+\nu)d}{2+p}}\lambda^\frac{-2p}{2+p}
\end{eqnarray}
with confidence at least $1-\frac\delta2$ (with respect to
$\rho^m$).

From (\ref{error decomposition}) to (\ref{bound s2}),
 we obtain
\begin{eqnarray*}
        &&\mathcal E(\pi_Mf_{{\bf z},\sigma,s\lambda,n})-\mathcal
        E(f_\rho)\\
        &\leq&
         C\left(\sigma^{2r}+\sigma^{2d}+\lambda/n
         +
         \frac{\log\frac4\delta}{3m}\right.\\
         &+&
        \left.\frac12\{\mathcal E(\pi_Mf_{{\bf z},\sigma,s\lambda,n})-\mathcal
        E(f_\rho)\}\right.\\
        &+&
        \left.\log\frac4\delta
        m^{-\frac{2}{2+p}}\sigma^{\frac{(p-2)(1+\nu)d}{2+p}}\lambda^\frac{-2p}{2+p}\right)
\end{eqnarray*}
holds with confidence at least $(1-\delta)\times
(1-2\exp\{-cn\sigma^{2d}\})$ (with respect to $\rho^m\times \mu^n)$.

Set $\sigma=m^{-\frac1{2r+d}+\varepsilon}$, $n=m^\frac{2d}{2r+d}$,
$\lambda=m^{-a}:=m^{-\frac{2r-d}{4r+2d}}$, $
\nu=\frac{\varepsilon}{2d(2r+d)}$ and
$$
           p=\frac{2d+2\varepsilon(2r+d)-2(1+\nu)+2(2r+d)\varepsilon(1+\nu)d}{(2r+d)(2a+d\varepsilon+\nu d\varepsilon-\varepsilon)+2r-(1+\nu)d}.
$$
Since $r\geq d/2$, it is easy to check that
$
            \nu>0$, and $0<p\leq 2$.
Then, we get
\begin{eqnarray*}
                &&\mathcal E(\pi_Mf_{{\bf z},\sigma,s\lambda,n})-\mathcal E(f_\rho)
        \leq
               Cm^{-\frac{2r}{2r+d}+\varepsilon}\log4\delta\\
               &+&
               m^{-\frac{2d}{2r+d}+\varepsilon}
               +
               \log4\delta
               m^{-\frac{2r+3d}{4r+d}}.
\end{eqnarray*}
Noting further that $r\leq d$, we obtain
$$
                \mathcal E(\pi_Mf_{{\bf z},\sigma,s\lambda,n})-\mathcal E(f_\rho)
        \leq
               Cm^{-\frac{2r}{2r+d}+\varepsilon}\log4\delta.
$$
Noticing the identity
$$
             E_{\rho^m} (\mathcal E(f_\rho)-\mathcal E(f_{{\bf
           z},\lambda,q}))=\int_0^\infty P^m\{\mathcal E(f_\rho)-\mathcal E(f_{{\bf
           z},\lambda,q})>\varepsilon\}d\varepsilon,
$$
direct computation yields the upper bound of (\ref{th2}). The lower
bound can be found in \cite[Chap.3]{Gyorfi2002}. This finishes the
proof of Theorem \ref{REGULARIZATION THEOREM}.
\end{IEEEproof}

\section{Conclusions}

The ELM-like learning provides a powerful computational burden
reduction  technique that adjusts only the output connections.
Numerous experiments and applications have demonstrated the
effectiveness and efficiency of ELM. The aim of our study is to
provide theoretical fundamentals of it. After analyzing the pros and
cons of ELM, we found that the theoretical performance of ELM
depends heavily on the activation function and randomness mechanism.
In the previous cousin paper \cite{Liu2013}, we have provided the
advantages of ELM in theory, that is, with appropriately selected
  activation function, ELM reduces the computation burden without
sacrificing the generalization capability in the expectation sense.
In this paper, we  discussed certain disadvantages of ELM. Via
rigorous proof, we found that ELM   suffered from both the
uncertainty
  and  generalization  degradation problem. Indeed, we proved
that, for the widely used Gaussian-type activation function, ELM
degraded the generalization capability. To facilitate the use of
ELM, some remedies of the aforementioned two problem are also
recommended. That is, multiple times trials can avoid the
uncertainty problem and
 the $l^2$ coefficient regularization
technique can essentially improve the generalization capability of
ELM. All these results reveal the essential characteristics of ELM
learning and give a feasible guidance concerning how to use ELM .

We conclude this paper with a crucial question about ELM learning.

\begin{question}\label{Q1}
As is shown in \cite{Liu2013} and the current paper, the performance
of ELM depends heavily on the activation function. For appropriately
selected activation function, ELM does not degrade the
generalization capability, while there also exists an activation
function such that the degradation exists. As it is impossible to
enumerate all the activation functions and study the generalization
capabilities of the corresponding ELM, we are asked for a  general
condition on the activation function, under which the corresponding
ELM    degrade (or doesn't degrade) the generalization capability.
In other words, we are interested in a criterion to classify the the
activation functions into two classes.  With the first class,   ELM
degrades the generalization capability and with the other class, ELM
does not degrades the generalization capability. We will keep
working on this interesting project, and report our progress in a
future publication.
\end{question}

\section*{Acknowledgement}
 The research was supported
by the National 973 Programming (2013CB329404), the Key Program of
National Natural Science Foundation of China (Grant No. 11131006),
and the National Natural Science Foundations of China (Grants No.
61075054).

\end{document}